# *On the Acceptability of Arguments in Preference-based Argumentation*

**Leila AMGOUD**    **Claudette CAYROL**
Institut de Recherche en Informatique de Toulouse (I. R. I. T.)
Université Paul Sabatier, 118 route de Narbonne, 31062 Toulouse Cedex (FRANCE)
E-mail: {amgoud, testemal}@irit.fr

**Abstract**

Argumentation is a promising model for reasoning with uncertain and inconsistent knowledge. The key concept of acceptability enables to differentiate arguments and defeaters: The certainty of a proposition can then be evaluated through the most acceptable arguments for that proposition. In this paper, we investigate different complementary points of view: an acceptability based on the existence of direct defeaters and an acceptability based on the existence of defenders. Pursuing previous work on preference-based argumentation principles, we enforce both points of view by taking into account preference orderings for comparing arguments. Our approach is illustrated in the context of reasoning with stratified knowledge bases.

## 1 INTRODUCTION

Various argument-based approaches to defeasible reasoning have been developed [LS89], [Vre91], [Pol92], [PL92], [SL92], [Dun93], [EFK93], [BCDLP93], [BDP93&95], [Cay95a&b] [EH95], [PS96]. Particularly, argumentation is a promising model for reasoning with inconsistent knowledge, based on the construction and the comparison of arguments. It may be also considered as a different method for handling uncertainty. The basic idea behind argumentation is that it should be possible to say more about the certainty of a particular fact than the certainty quantified with a degree in [0, 1]. In particular, it should be possible to assess the reason why a fact holds, under the form of arguments, and combine these arguments for the certainty evaluation. Indeed, the process of combination may be viewed as a kind of reasoning about the arguments in order to determine the most acceptable of them. For that purpose, we can take into account the existence of arguments in favor of, or against a given fact as well as preference orderings for comparing pairs of arguments.

The main approaches which have been developed for reasoning within an argumentation system rely on the idea of differentiating arguments with a notion of acceptability. Two kinds of acceptability have been proposed:

- An acceptability level is assigned to a given argument depending on the existence of direct defeaters, or defeaters. That leads to the concept of acceptability class introduced by [EFK93] and [EH95].
- Acceptability with respect to a rational agent relies upon a notion of defense [Dun93&95]: The set of all the arguments that a rational agent may accept must defend itself against any defeater. In the following, that point of view will be referred to as *joint acceptability*.

These notions of acceptability have been applied in a framework where defeat relations between arguments are based on the comparison of the structure of the arguments (logical criteria). For instance, an argument supporting a given fact defeats an argument supporting the opposite fact. In that restricted framework and in the particular case of inconsistent knowledge bases handling, close correspondences have been established between the two notions [Cay95a]. These two kinds of acceptability have been proved complementary.

Our purpose is to investigate the notion of acceptability by taking into account both points of view (individual and joint acceptability) and by combining purely logical criteria and preference orderings for comparing arguments.

In previous works, [Cay95b], we have proposed a methodological approach to the integration of preference orderings into acceptability classes. In [AC97], we have proposed two categories of preference-based argumentation frameworks with new acceptability classes. It is convenient to consider one such class as a set of arguments which can defend themselves against all attacks. This notion of *individual defense* is modeled via the preferences between arguments. Unfortunately, these frameworks enable to take into account only direct defeaters since they consider only one level of defeat. To illustrate that point, let us consider the following example: Let A, B, C be three arguments such that B defeats A, C defeats B, B is preferred to A and C is preferred to B. In that example the argument A does not defend itself against B. Then A will not belong to the acceptability class. However, B is itself defeated by C. As C defeats B and is preferred to B, C defends A. So, taking into account two levels of defeat, A should be acceptable.
The idea we enforce is the following one: we accept an argument if it is not defeated or if it *defends itself* against all attacks (for instance because it is preferred to its defeaters) or if it is defended by other arguments. This latter case corresponds to a *joint defense*, in the sense of Dung's work. We intend to apply Dung's general construction for preference-based argumentation.

This paper is organized as follows: section 2 is devoted to argumentation frameworks based on acceptability classes. Firstly, we recall the basic argumentation frameworks as proposed in [EFK93], in the particular case of handling inconsistent knowledge bases. Then, we show how introducing preferences into acceptability classes enables to define simple notion of individual defense. In section 3, we apply the general framework proposed by Dung to preference-based argumentation. We show that our proposal is convenient for coping with different kinds of acceptability. We give, in section 4, some preliminary results on the relationships between argumentation-based reasoning and prioritized-coherence based inference schemas.

## 2  ARGUMENTATION FRAMEWORKS BASED ON ACCEPTABILITY CLASSES

### 2.1  BASIC DEFINITIONS

**Definition 1.** An *argumentation framework* is a pair $<\mathcal{A}, \mathcal{R}>$ where $\mathcal{A}$ is a set of arguments and $\mathcal{R}$ is a binary relation representing a defeat relationship between arguments, i.e. $\mathcal{R} \subseteq \mathcal{A} \times \mathcal{A}$. $(A, B) \in \mathcal{R}$ or equivalently "A $\mathcal{R}$ B" means that the argument A defeats the argument B. Here, an argument is an abstract entity whose role is only determined by its relation to other arguments.

**Definition 2.** The *acceptability class* w.r.t the argumentation framework $<\mathcal{A}, \mathcal{R}>$ is $\{A \in \mathcal{A} \mid$ there does not exist $B \in \mathcal{A}$ such that $B \mathcal{R} A\}$ and will be denoted by $C_{\mathcal{R}}$ (if no ambiguity on $\mathcal{A}$).

In other words, the acceptability class contains the arguments which are not defeated by another argument. On the same set of arguments, for each defeat relation an acceptability class can be defined.

**Example 1.** Let $<\mathcal{A}, \mathcal{R}>$ be the argumentation framework defined by $\mathcal{A} = \{A, B, C, D\}$ and $\mathcal{R} = \{(C, D), (D, C)\}$. The arguments A and B are not defeated so the acceptability class is $C_{\mathcal{R}} = \{A, B\}$.

Examples of such systems are the ones proposed by Elvang & al. in [EFK93] for handling inconsistency in knowledge bases. The arguments are built from an inconsistent knowledge base $\Sigma = (E, K)$. The sets E and K contain formulas of a propositional language L. K represents a core of knowledge and is assumed consistent. Contrastedly, formulas of E represent defeasible pieces of knowledge, or beliefs. So $K \cup E$ may be inconsistent. The following definition of argument is used:

**Definition 3.** An *argument* of E in the context K is a pair (H, h), where h is a formula of the language L and H a subbase of E satisfying: i) $K \cup H$ is consistent, ii) $K \cup H \vdash h$, iii) H is minimal (no strict subset of H satisfies i and ii). H is called the support and h the conclusion of the argument.

We denote by $\mathcal{A}(\Sigma)$ the set of all the arguments which are constructed from a knowledge base $\Sigma$. The following defeat relations are borrowed from [EFK93].

**Definition 4.** Let (H, h) and (H', h') be two arguments of $\mathcal{A}(\Sigma)$.
(H, h) *rebuts* (H', h') iff $h \equiv^1 \neg h'$. This means that an argument is rebutted if there exists an argument for the negated conclusion.
(H, h) *undercuts* (H', h') iff for some $k \in H'$, $h \equiv \neg k$. An argument is undercut if there exists an argument against one element of its support.

Two acceptability classes denoted respectively by $C_{\text{Rebut}}$ and $C_{\text{Undercut}}$ are associated with these relations.
$C_{\text{Rebut}}$ is the set of arguments of $\Sigma$ which are not rebutted by some argument of $\Sigma$.
$C_{\text{Undercut}}$ is the set of arguments of $\Sigma$ which are not undercut by some argument of $\Sigma$
It can be proved that: $C_{\text{Undercut}} \subseteq C_{\text{Rebut}}$. See [Cay95a] for a discussion about other defeat relations on $\mathcal{A}(\Sigma)$.

Note that a similar methodology for defining the concept of "defeat" is used in [PS95&96] with the same terminology but with a different structure of the arguments. In [PS95&96], an argument is a sequence of chained implicative rules. Each rule has a consequent part (consisting of one literal) and an antecedent part (consisting of a conjunction of literals). The consequent of each rule in a given argument is considered as a conclusion of that argument.

### 2.2  PREFERENCE-BASED ARGUMENTATION FRAMEWORKS

The notion of acceptability has been most often defined purely on the basis of other constructible arguments. But other criteria may be taken into account for comparing arguments. In the case of knowledge bases, for instance, specificity [SL92], or explicit priorities on the beliefs can be taken into account. More generally, preference relations can be used for comparing arguments. Indeed, recent works (for instance [Gro91], [CRS93], [Bre94], [CLS95]) have proved that preference relations allow for more sophisticated and more appropriate handling of conflict resolution and uncertain knowledge. (See [ACL96] for a more general discussion on preferences and arguments).
To enforce the concept of acceptability used in basic argumentation frameworks, we have introduced in [AC97] preference orderings into the definition of acceptability classes. Instead of keeping only the arguments which are not defeated, we accept also the arguments which are preferred to their defeaters. We say that such an argument defends itself against all attacks.

---

$^1 \equiv$ denotes logical equivalence

**Definition 5.** An argument A *defends itself* against an argument B which defeats A iff A is preferred to B.
In that sense, preference orderings model a notion of defense.

**Definition 6.** A *preference-based argumentation framework* is a triplet $<\mathcal{A}, \mathcal{R}, Pref>$ where Pref is a partial preordering (reflexive and transitive binary relation) on $\mathcal{A} \times \mathcal{A}$.

**Notation:** $>>^{Pref}$ is the strict partial ordering associated with Pref.

In a preference-based argumentation framework, the acceptability concept can be weakened:

**Definition 7.** The *acceptability class* w.r.t the argumentation framework $<\mathcal{A}, \mathcal{R}, Pref>$ is $\{A \in \mathcal{A} \mid \forall B \in \mathcal{A}$ if $B \mathcal{R} A$ then $A >>^{Pref} B\}$ and will be denoted by $C_{\mathcal{R},Pref}$. In other words, an acceptability class contains all the arguments which defend themselves against any attack.

In example 1, if we suppose that $C >>^{Pref} D$, then the acceptability class corresponding to the argumentation framework $<\mathcal{A}, \mathcal{R}, Pref>$ is $C_{\mathcal{R},Pref} = \{A, B, C\}$.

In [AC97], from the two defeat relations "rebut" and "undercut" of [EFK93], we have developed two categories of preference-based argumentation frameworks. In each one, several systems can be discussed according to different definitions for the relation Pref. With respect to $<\mathcal{A}(\Sigma), Rebut, Pref>$, the acceptability class called $C_{Rebut,Pref}$ gathers all the arguments which defend themselves against their rebutting arguments. The second category is defined by the triplet $<\mathcal{A}(\Sigma), Undercut, Pref>$. The corresponding acceptability class, called $C_{Undercut,Pref}$, gathers all the arguments which defend themselves against their undercutting arguments. (See [AC97] and [AC98a] for other acceptability classes and a complete study on these classes).

The following examples will illustrate our claim: we consider the preference relation proposed in [BDP93] in a possibilistic context and based on the certainty level. In that case, the knowledge base $\Sigma$ is supposed to be stratified in $\Sigma = \Sigma_1 \cup ... \cup \Sigma_n$ such that beliefs in $\Sigma_i$ have the same certainty level and are more reliable than beliefs in $\Sigma_j$ where $j > i$. The certainty level of a non-empty subset H of $\Sigma$ is the highest number of a layer (i.e. the lower layer) met by H.

**Definition 8.** The certainty level of an argument (H, h) of $\mathcal{A}(\Sigma)$ is exactly the certainty level of the set H.

**Example 2.** Let $K = \emptyset$, $E = \Sigma = \Sigma_1 \cup ... \cup \Sigma_3$ be a knowledge base with $\Sigma_1 = \{a, \neg a\}$, $\Sigma_2 = \{a \rightarrow b\}$ and $\Sigma_3 = \{\neg b\}$. The certainty level of the argument $(\{a, a \rightarrow b\}, b)$ is 2 whereas the certainty level of the argument $(\{\neg b\}, \neg b)$ is 3, then the argument $(\{a, a \rightarrow b\}, b)$ is preferred to the argument $(\{\neg b\}, \neg b)$ because it has a higher certainty level. We say that $(\{a, a \rightarrow b\}, b)$ defends itself against the unique rebutting argument and consequently it belongs to the acceptability class $C_{Rebut,Pref}$. However, it does not belong to $C_{Undercut,Pref}$ because it can't defend itself against the undercutting argument $(\{\neg a\}, \neg a)$ whose certainty level is 1.

The framework using the classes $C_{\mathcal{R},Pref}$ is however too restricted since it discards arguments which appear acceptable. Intuitively, if an argument A is less preferred than its defeater B then it is weakened. But the defeater B itself may be weakened by another argument C which defeats B and is preferred to B. In this latter case we would like to accept A. To illustrate this point, let us take the following example.

**Example 3.** Let $K = \emptyset$, $E = \Sigma = \Sigma_1 \cup ... \cup \Sigma_4$ with $\Sigma_1 = \{x, \neg r\}$, $\Sigma_2 = \{x \rightarrow t\}$, $\Sigma_3 = \{t \rightarrow r\}$ and $\Sigma_4 = \{\neg r \rightarrow p\}$. The argument $(\{x, x \rightarrow t, t \rightarrow r\}, r)$ undercuts the argument $(\{\neg r, \neg r \rightarrow p\}, p)$ and $(\{x, x \rightarrow t, t \rightarrow r\}, r)$ is preferred to $(\{\neg r, \neg r \rightarrow p\}, p)$. So $(\{\neg r, \neg r \rightarrow p\}, p)$ does not defend itself and consequently it does not belong to the acceptability class $C_{Undercut,Pref}$. However, the argument $(\{x, x \rightarrow t, t \rightarrow r\}, r)$ is itself undercut by a preferred one, $(\{x, x \rightarrow t, \neg r\}, \neg(t \rightarrow r))$, and it can't defend itself. In this case we claim that the argument $(\{x, x \rightarrow t, \neg r\}, \neg(t \rightarrow r))$ defends the argument $(\{\neg r, \neg r \rightarrow p\}, p)$ against $(\{x, x \rightarrow t, t \rightarrow r\}, r)$. So the proposition "p" may be concluded from $\Sigma$.

## 3 JOINT ACCEPTABILITY

### 3.1 DUNG'S GENERAL ARGUMENTATION FRAMEWORK

Inspired by earlier work of Bondarenko, Kakas, Kowalski and Toni, [Dun93&95] has proposed a very abstract and general argument-based framework $<\mathcal{A}, Att>$ where Att denotes any attack relation between arguments. An up-to-date technical survey of this approach is [BDKT95]. The basic idea is that a fact holds if an argument supporting this fact can be defended against all the defeaters. For a rational agent, an argument A is acceptable if he can defend A against all attacks on A. All the arguments acceptable for a rational agent will be gathered in a so-called extension. An extension must satisfy a coherence requirement as defined by:

**Definition 9.** A set $S \subseteq \mathcal{A}$ of arguments is *conflict-free* iff no argument in S is attacked by another argument of S.

**Definition 10.** An argument A is *defended* by a set S of arguments (or S *defends* A) iff $\forall$ B $\in$ $\mathcal{A}$, if B Att A then $\exists$ C $\in$ S such that C Att B.

The joint acceptability requirement characterizes the principle: An agent can defend the arguments he accepts and accepts every argument he can defend. An extension satisfying such a requirement is defined by a fixpoint construction as follows:

**Definition 11.** Let <$\mathcal{A}$, Att> be an argumentation framework. Let us define a function $\mathcal{F}$ as:
$$\mathcal{F}: 2^{\mathcal{A}} \to 2^{\mathcal{A}}$$
$$S \to \mathcal{F}(S) = \{A \in \mathcal{A} | \text{ A is defended by S}\}$$

**Definition 12.** A conflict-free set of arguments S is a *complete extension* iff S is a fixed point of $\mathcal{F}$.

Another kind of extension may be defined with a stability requirement.

**Definition 13.** Let <$\mathcal{A}$, Att> be an argumentation framework. Let us define a function $\mathcal{G}$ as:
$$\mathcal{G}: 2^{\mathcal{A}} \to 2^{\mathcal{A}}$$
$$S \to \mathcal{G}(S) = \{A \in \mathcal{A} | \text{ A is not attacked (w.r.t the relation Att) by S}\}$$

**Definition 14.** A conflict-free set of arguments S is a stable extension iff S is a fixed point of $\mathcal{G}$.

The following characterization can be easily proved:

**Property 1.** A conflict-free set of arguments S is a *stable extension* iff each argument which is not in S is attacked by some argument in S.

Note that a similar definition appears in [Pol92] and a variant of Dung's formalization leads to the set of justified arguments in [PS95&96].

The following results are proved in [Dun95]:
- Let S, S' $\subseteq$ $\mathcal{A}$. If S $\subseteq$ S' then $\mathcal{F}(S) \subseteq \mathcal{F}(S')$.
- Let S $\subseteq$ $\mathcal{A}$. If S is conflict-free then $\mathcal{F}(S)$ is also conflict-free.
- Every stable extension is a complete one, but the converse does not hold.
  *Proof: Take $\mathcal{A}$ = {A} and Att = {(A, A)}. The empty set is a complete extension but it is not a stable one.*
- Every argumentation framework has at least one complete extension.

Moreover, we have proved [AC98b]:
**Proposition 1.**
- Let S, S' $\subseteq$ $\mathcal{A}$. If S $\subseteq$ S' then $\mathcal{G}(S') \subseteq \mathcal{G}(S)$.
- Let S $\subseteq$ $\mathcal{A}$. S is conflict-free iff S $\subseteq$ $\mathcal{G}(S)$.
- Let S $\subseteq$ $\mathcal{A}$. If S is a stable extension then S is maximal (for set inclusion) among the sets which are conflict-free.
- Let S $\subseteq$ $\mathcal{A}$. $\mathcal{F}(S) = \mathcal{G} \circ \mathcal{G}(S)$.

In example 1, the two sets {A, B, C} and {A, B, D} are complete extensions and also stable extensions. {A, B} is a complete but not stable extension.

### 3.2 DEFENSE IN PREFERENCE-BASED ARGUMENTATION

In the context of preference-based argumentation framework <$\mathcal{A}$, $\mathcal{R}$, Pref>, the general schema proposed by Dung will enable us to bypass the limit encountered with the classes $C_{\mathcal{R},\text{Pref}}$ in section 2.2.

Here, $\mathcal{R}$ represents a defeat relationship based on purely logical properties (such as "rebut" or "undercut", for instance) and Pref is a preference relation based on meta-knowledge which cannot be extracted from the arguments themselves.

**Definition 15.** Let A, B be two arguments. We define B Att A iff B $\mathcal{R}$ A and not (A $>>^{\text{Pref}}$ B).

In other words, an argument A is attacked if it is defeated by another argument in the sense of $\mathcal{R}$ and it cannot defend itself using Pref. Note that in this case, a set of arguments S is conflict-free if each argument in S defends itself against its defeaters which are in S.

Let <$\mathcal{A}$, $\mathcal{R}$, Pref> be an argumentation framework. All the definitions 9, 10, 11, 12, 13, 14 apply in the framework <$\mathcal{A}$, Att> defined through definition 15. More precisely, we have:

$\mathcal{F}(S)$ = {A $\in$ $\mathcal{A}$| if B $\mathcal{R}$ A and not (A $>>^{\text{Pref}}$ B), then there exists C $\in$ S such that C $\mathcal{R}$ B and not (B $>>^{\text{Pref}}$ C)}

$\mathcal{G}(S)$ = {A $\in$ $\mathcal{A}$| there does not exist B $\in$ S such that B $\mathcal{R}$ A and not (A $>>^{\text{Pref}}$ B)}, or equivalently $\mathcal{G}(S)$ = { A $\in$ $\mathcal{A}$| if B $\in$ S and B $\mathcal{R}$ A then A $>>^{\text{Pref}}$ B}.

**Example 4.** Let <$\mathcal{A}$, $\mathcal{R}$, Pref> be an argumentation framework where $\mathcal{A}$ = {A, B, C}, $\mathcal{R}$ = {(A, B), (B, C)}. Let us suppose that B $>>^{\text{Pref}}$ A and C $>>^{\text{Pref}}$ B then Att = $\emptyset$. The set S = {A, B, C} is conflict-free and it is a complete extension.

Our purpose is now to characterize the complete and the stable extensions w.r.t the class $C_{\mathcal{R},\text{Pref}}$. All the proofs can be found in [AC98b].

**Proposition 2.** Let <$\mathcal{A}$, $\mathcal{R}$, Pref> be an argumentation framework.
- $\mathcal{G}(\emptyset) = \mathcal{A}$.
- $\mathcal{G}(\mathcal{A}) = C_{\mathcal{R},\text{Pref}}$.

For the proof see [AC98b].
As a direct consequence of these results we have:
- The set $C_{\mathcal{R},\text{Pref}}$ is conflict-free.
  *Proof: $C_{\mathcal{R},\text{Pref}} \subseteq \mathcal{A}$ then $\mathcal{G}(\mathcal{A}) = C_{\mathcal{R},\text{Pref}} \subseteq \mathcal{G}(C_{\mathcal{R},\text{Pref}})$*

- $\mathcal{F}(\emptyset) = C_{\mathcal{R},\text{Pref}}$.

From classical results in fixed-point theory, the least fixed point of $\mathcal{F}$ can be computed by iteratively applying $\mathcal{F}$ to $\emptyset$, provided that $\mathcal{F}$ is continuous on the complete lattice $2^{\mathcal{A}}$. Moreover, Dung has proved that $\mathcal{F}$ is continuous if the argumentation framework is finitary (that is : for each argument A, there are only finitely many arguments which attack A).

All the following results hold in the case of a finitary argumentation framework $<\mathcal{A}, \mathcal{R}, \text{Pref}>$. Proofs can be found in the full report [AC98b].

**Theorem 1.** Let $\underline{S}$ denote the least fixed point of the function $\mathcal{F}$:
$\underline{S} = \cup \mathcal{F}^i(\emptyset), i \geq 0 = C_{\mathcal{R},\text{Pref}} \cup [\cup \mathcal{F}^i(C_{\mathcal{R},\text{Pref}}), i \geq 1]$.

**Proposition 3.** The subset $\underline{S}$ is conflict-free.

**Proposition 4.**
- The least (for set-inclusion) complete extension is $\underline{S}$.
- Each complete (resp. stable) extension contains the subset $\underline{S}$.

The above result is not surprising since our aim was to keep the arguments which can defend themselves against all attacks on them and to add those which are defended by other arguments.

**Proposition 5.**
- Let $S \subseteq \mathcal{A}$. If S is a fixed point of $\mathcal{F}$ then $\mathcal{G}(S)$ is also a fixed point of $\mathcal{F}$.
- $\mathcal{G}(\underline{S})$ is the *greatest (for set-inclusion) fixed point* of the function $\mathcal{F}$.

So, each complete extension S satisfies: $\underline{S} \subseteq S \subseteq \mathcal{G}(\underline{S})$. Moreover, we have the equivalence:
$\underline{S}$ is the *unique complete extension* iff $\mathcal{G}(\underline{S})$ is conflict-free.

In the particular context of reasoning with stratified knowledge bases, the argumentation frameworks $<\mathcal{A}(\Sigma)$, Rebut, Pref> and $<\mathcal{A}(\Sigma)$, Undercut, Pref> (introduced in section 2.2) are finitary. It is due to the use of a propositional language and finite knowledge bases. Then, all the above results hold.

**Example 3. (Continuation)** We have shown that the argument $(\{\neg r, \neg r \rightarrow p\}, p)$ does not belong to the acceptability class $C_{\text{Undercut,Pref}}$ because it cannot defend itself against the argument $(\{x, x \rightarrow t, t \rightarrow r\}, r)$ which undercuts it. However, $(\{x, x \rightarrow t, t \rightarrow r\}, r)$ is itself undercut by $(\{x, x \rightarrow t, \neg r\}, \neg(t \rightarrow r))$, and the argument $(\{x, x \rightarrow t, \neg r\}, \neg(t \rightarrow r))$ is preferred to $(\{x, x \rightarrow t, t \rightarrow r\}, r)$. Then we say that the argument $(\{x, x \rightarrow t, \neg r\}, \neg(t \rightarrow r))$ defends the argument $(\{\neg r, \neg r \rightarrow p\}, p)$ and $(\{\neg r, \neg r \rightarrow p\}, p) \in \mathcal{F}^1$

$(C_{\text{Undercut,Pref}})$. As a consequence, the argument $(\{\neg r, \neg r \rightarrow p\}, p)$ belongs to the least fixed point of the function $\mathcal{F}$ and also to every complete or stable extension.

## 4 RELATED ISSUES

The coherence notion proposed in the argumentation frameworks $<\mathcal{A}, \mathcal{R}, \text{Pref}>$ may appear too permissive since $\mathcal{R}$-conflicting arguments may be present in the same extension (two arguments A and B are said $\mathcal{R}$-conflicting iff A defeats B in the sense of the relation $\mathcal{R}$). The coherence may be enforced by a new definition:

**Definition 16.** A set $S \subseteq \mathcal{A}$ of arguments is *conflict-free* iff no argument in the set is defeated in the sense of $\mathcal{R}$ by another argument in S. In other words there doesn't exist A and B in S such that A $\mathcal{R}$ B.

That restricted coherence property is satisfied by the acceptability classes of the frameworks $<\mathcal{A}(\Sigma)$, Rebut, Pref> and $<\mathcal{A}(\Sigma)$, Undercut, Pref>. Indeed, we have:

**Proposition 6.** If $(H, h) \in C_{\text{Undercut,Pref}}$ then there does not exist $(H', h')$ such that $(H', h')$ undercuts $(H, h)$ and $(H', h') \in C_{\text{Undercut,Pref}}$. (The same result holds in the framework $<\mathcal{A}(\Sigma)$, Rebut, Pref>).

Moreover, the above proposition enables to establish correspondences between two main approaches of reasoning with inconsistent knowledge bases: argumentation-based approaches and coherence-based approaches. In the case of flat bases (no preference is expressed on the beliefs), we have proved that the consequence relations defined through the argumentation frameworks $<\mathcal{A}(\Sigma)$, Undercut> and $<\mathcal{A}(\Sigma)$, rebut> correspond exactly to the ones proposed using maximal consistent subbases in the coherence-based approaches [Cay95a]. For instance, in the framework $<\mathcal{A}(\Sigma)$, Undercut>, the stable extensions are exactly the sets Arg(T) where T is a maximal (for set-inclusion) K-consistent subset of $\Sigma$ and Arg(T) denotes the set of all the arguments whose support is included in T. Similarly, preference-based argumentation can be connected with prioritized coherence-based inference schemas (See [CLS95] for a thorough presentation of these inference schemas). As in example 3, $K = \emptyset$ and $\Sigma$ is supposed to be stratified in $\Sigma_1 \cup \ldots \cup \Sigma_n$. Pref is defined by the certainty level.

**Definition 17.** A consistent subbase $S = S_1 \cup \ldots \cup S_n$ is an INCL-preferred subbase of $\Sigma$ if and only if $\forall j=1..n$, $S_1 \cup \ldots \cup S_j$ is a maximal (for set-inclusion) consistent subbase of $\Sigma_1 \cup \ldots \cup \Sigma_j$. INCL($\Sigma$) denotes the set of INCL-preferred subbases of $\Sigma$ and $\cap$ INCL($\Sigma$) denotes the intersection of the INCL-preferred subbases of $\Sigma$. B being a subset of $\mathcal{A}(\Sigma)$, Supp(B) denotes the union of the supports of arguments of B. We have proved [AC97&AC98B]:

**Proposition 7.**
- If $T \in INCL(\Sigma)$ then $Arg(T)$ is a stable extension of the argumentation framework $<\mathcal{A}(\Sigma), Undercut, Pref>$.
- $C_{Undercut,Pref} \subseteq \cap\, INCL(\Sigma)$.

It follows that:
- $C_{Undercut,Pref} \subseteq$ each stable extension of the argumentation framework $<\mathcal{A}(\Sigma), Undercut, Pref>$.

Owing to the above results, we think that the least complete extension is the key concept for connecting argumentation-based reasoning with inference schemas based on coherence restauration.

## 5 CONCLUSION

The work reported here concerns the acceptability of arguments in preference-based argumentation. Our first contribution is to take into account preference relations between arguments in order to select the most acceptable arguments. The use of preferences enables to model a notion of individual defense. In previous work, we have proposed an abstract and general argumentation framework where an argument is acceptable only if it defends itself against each defeater. We have studied two applications in the context of reasoning with inconsistent/uncertain stratified knowledge bases. However, this kind of acceptability considers only direct defeaters of a given argument. Our second contribution is to enforce the concept of joint acceptability: an argument may be acceptable by a rational agent if it is defended by other arguments of the same agent. As this notion of defense has been proposed in [Dun93], we have extended our general framework by applying the Dung's schema. Moreover, we have obtained preliminary results concerning the relationship between preference-based argumentation and prioritized coherence-based approaches to the handling of knowledge bases. We are now working in that direction. Our purpose is to develop a unifying framework for inference schemas coping with inconsistency and uncertainty.

**ACKOWLEDGEMENTS**

We would like to thank Laurent Garcia and the referees for their detailed comments which helped us to improve this paper.